\documentclass[10pt,twocolumn,letterpaper]{article}
\usepackage[toc,title,page]{appendix}
\usepackage[accsupp]{axessibility}
\usepackage{iccv}
\usepackage{times}
\usepackage{epsfig}
\usepackage{graphicx}
\usepackage{amsmath}
\usepackage{amssymb}
\usepackage{arydshln}
\usepackage{booktabs}
\usepackage{bbding}
\usepackage{algorithm}
\usepackage{algpseudocode}
\usepackage{enumitem}
\usepackage{multirow}
\usepackage{capt-of}
\usepackage{bbm}
\usepackage{dsfont}
\usepackage[breaklinks=true,bookmarks=false]{hyperref}

\iccvfinalcopy 

\ificcvfinal\fi

\begin{document}

\title{Object-Centric Multiple Object Tracking}

\author{Zixu Zhao$^{1}$ \quad Jiaze Wang$^{2}$\thanks{Work completed during internship at AWS Shanghai AI Lab.} \quad Max Horn$^{1}$ \quad  Yizhuo Ding$^{3*}$ \quad Tong He$^{1}$ \quad Zechen Bai$^{1}$ \\
Dominik Zietlow$^{1}$ \quad Carl-Johann Simon-Gabriel$^{1}$ \quad Bing Shuai$^{1}$ \quad Zhuowen Tu$^{1}$ \quad Thomas Brox$^{1}$\\ 
 Bernt Schiele$^{1}$ \quad Yanwei Fu$^{3}$ \quad Francesco Locatello$^{1}$ \quad Zheng Zhang$^{1}$\thanks{Corresponding author.} \quad Tianjun Xiao$^{1}$\\
$^{1}$ Amazon Web Services \quad$^{2}$ The Chinese University of Hong Kong \quad  $^{3}$ Fudan University\\
{\tt\small \{zhaozixu, jiazew, yizhuodi, htong, baizeche, bshuai, ztu, zhaz, tianjux\}@amazon.com}\\
{\tt\small \{hornmax, zietld, cjsg, brox, bschiel, locatelf\}@amazon.de, yanweifu@fudan.edu.cn}\\
}


\maketitle
\ificcvfinal\thispagestyle{empty}\fi

\begin{abstract}
Unsupervised object-centric learning methods allow the partitioning of scenes into entities without additional localization information and are excellent candidates for reducing the annotation burden of multiple-object tracking (MOT) pipelines. Unfortunately, they lack two key properties: objects are often split into parts and are not consistently tracked over time. In fact, state-of-the-art models achieve pixel-level accuracy and temporal consistency by relying on supervised object detection with additional ID labels for the association through time.
This paper proposes a video object-centric model for MOT. It consists of 
an index-merge module that adapts the object-centric slots into detection outputs and an object memory module that builds complete object prototypes to handle occlusions.  Benefited from object-centric learning, we only require sparse detection labels ($0\%$-$6.25\%$) for object localization and feature binding. Relying on our self-supervised Expectation-Maximization-inspired loss for object association, our approach requires no ID labels.
Our experiments significantly narrow the gap between the existing object-centric model and the fully supervised state-of-the-art and outperform several unsupervised trackers. Code is available at 
\href{https://github.com/amazon-science/object-centric-multiple-object-tracking}{https://github.com/amazon-science/object-centric-multiple-object-tracking}.
\end{abstract}

\section{Introduction}


Visual indexing theory~\cite{pylyshyn1989role} proposes a psychological mechanism that includes a set of indexes that can be associated with an object in the environment. 
Each index retains its association with an object, even when that object moves, interacts with other objects, or becomes partially occluded. 
This theory was originally developed in the cognitive sciences, however, a very similar principle lies at the heart of object-centric representation learning.
By learning object-level representations, we can develop models inferring object relations~\cite{mambelli2022compositional,wu2022slotformer,yoon2023investigation} and even their causal structure~\cite{liu2023causal,mansouri2022object}. 
Additionally, object-centric representations have shown to be more robust \cite{dittadi2022generalization}, allow for combinatorial generalization \cite{liu2023causal}, and are beneficial for various downstream applications \cite{wu2022slotformer}.
Since causal relations often unfold in time, it is only logical to combine object-centric learning (OCL) with temporal dynamics modeling, where consistent object representations are necessary.

\begin{figure}[!t]
	\centering
	\includegraphics[width=82mm]{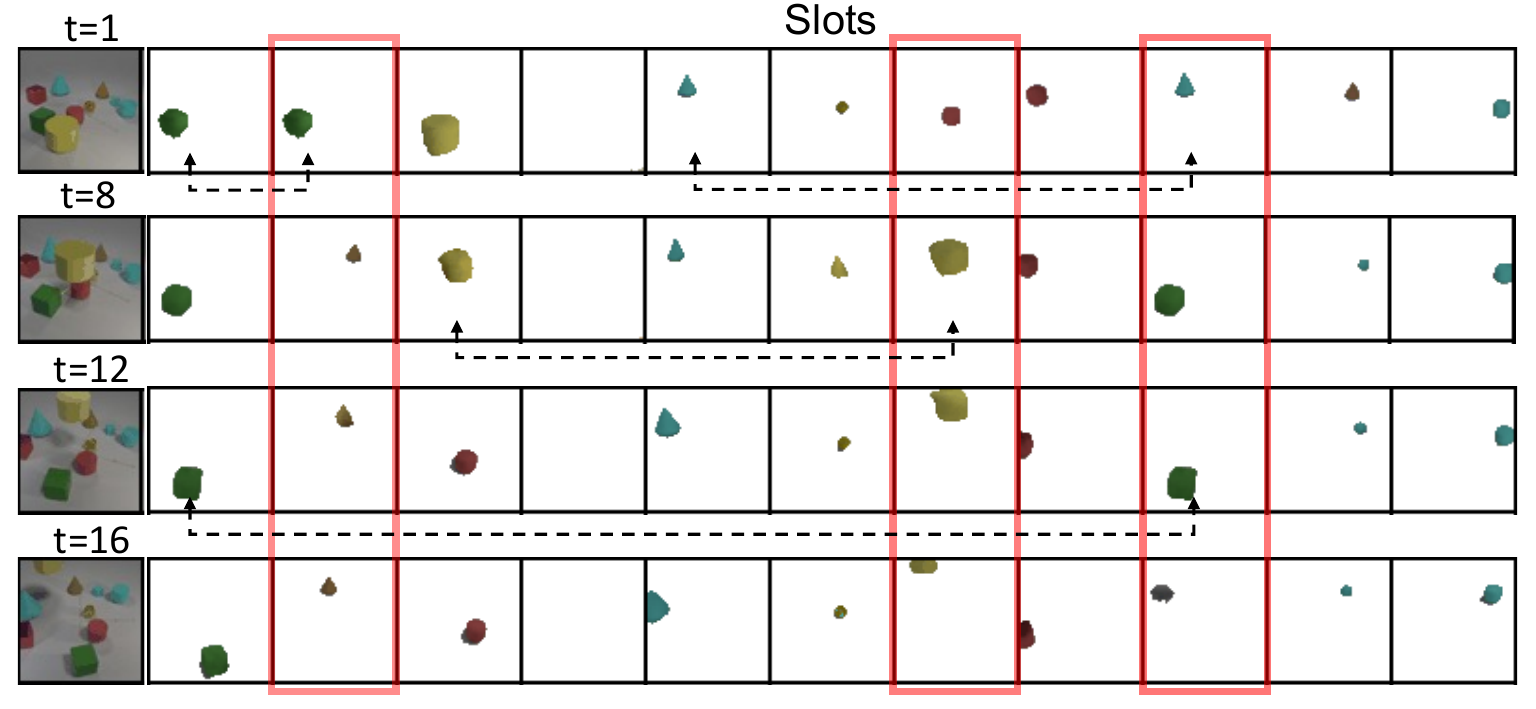}
 \vspace{+0.3cm}
\caption{\textbf{Temporal inconsistency and part-whole split of object-centric representations. } We visualize a video object-centric model SAVi~\cite{kipf2022conditional} that groups objects into a set of slots without labels. Common issues include that there exist many duplicate slots that capture the same object or its parts (dashed arrows),  and the slots fail to track objects consistently over time (red boxes). 
 }
\label{fig:teaser}
\end{figure}

Multiple object tracking (MOT) is a computer-vision problem that resembles visual indexing theory.
MOT aims at localizing a set of objects while following their trajectories over time so that the same object keeps the same identity in the entire video stream.
The dominant MOT methods follow the detect-to-track paradigm: First, employ an object detector to localize objects in each frame, then perform association on detected objects between adjacent frames to get tracklets.
The development of state-of-the-art MOT pipelines usually requires large amounts of detection labels for the objects we are interested in, as well as video datasets with object ID labels to train the association module.
Consequently, such approaches are label intense and do not generalize well in open-world scenarios.


Unsupervised object-centric representation learning tackles the object discovery and binding problem in visual data without additional supervision \cite{seitzer2022bridging}. 
Recent work, such as SAVi~\cite{kipf2022conditional} and STEVE~\cite{singh2022simple}, extended such models to the video domain, which hints at possible applications to MOT. 
However, existing approaches are primarily evaluated without heavy punishment if slots exchange ``ownerships'' of pixels and rather rely on clustering similarity metrics such as FG-ARI~\cite{kipf2022conditional}. An object may appear in different slots across time (a.k.a ID switch issue),  which hinders downstream applications of OCL models, especially when \emph{directional relationship} among objects and their dynamics must be reasoned upon (e.g., who acts upon whom). Additionally, the part-whole issues are not fully explored, allowing slots to only track parts of an object. Figure~\ref{fig:teaser} visualizes the two problems of OCL models that are developed orthogonally with respect to MOT downstream tasks, leading to a significant gap with the state-of-the-art fully supervised MOT methods.
Scalability challenges of unsupervised OCL methods only accentuate this gap.

In this work, we take steps to bridge the gap between object-centric learning and fully-supervised multiple object tracking pipelines. Our design focuses on improving OCL framework on two key issues: 1) track objects as a whole, and 2) track objects consistently over time. For these, we insert a memory model to consolidate slots into memory buffers (to solve the part-whole problem) and roll past representations of the memory forward (to improve temporal consistency). 
Overall, our model provides a label-efficient alternative to the otherwise costly MOT pipelines that rely on detection and ID labels.
Our contributions can be summarized as follows:
\begin{enumerate}[label=(\arabic*)]
    \item We develop a video object-centric model that can be applied to MOT task with very few detection labels ($0\,\%$-$6.25\,\%$) and no ID labels.
    \item OC-MOT leverages an unsupervised memory to predict completed future object states even if occlusion happens. Besides, the index-merge module can tackle the part-whole and duplication issues specific to OC models. The two cross-attention design is simple but nontrivial, serving as the “index” and “merge” functions with their key and query being bi-directional.
    \item We are the first to introduce the object-centric representations to MOT that are versatile enough in a way of supporting all the association, rolling-out, and merging functions, and can be trained with low labeling cost.

\end{enumerate}



\section{Related Works}

\begin{figure*}[t]
	\centering
	\includegraphics[width=\textwidth]{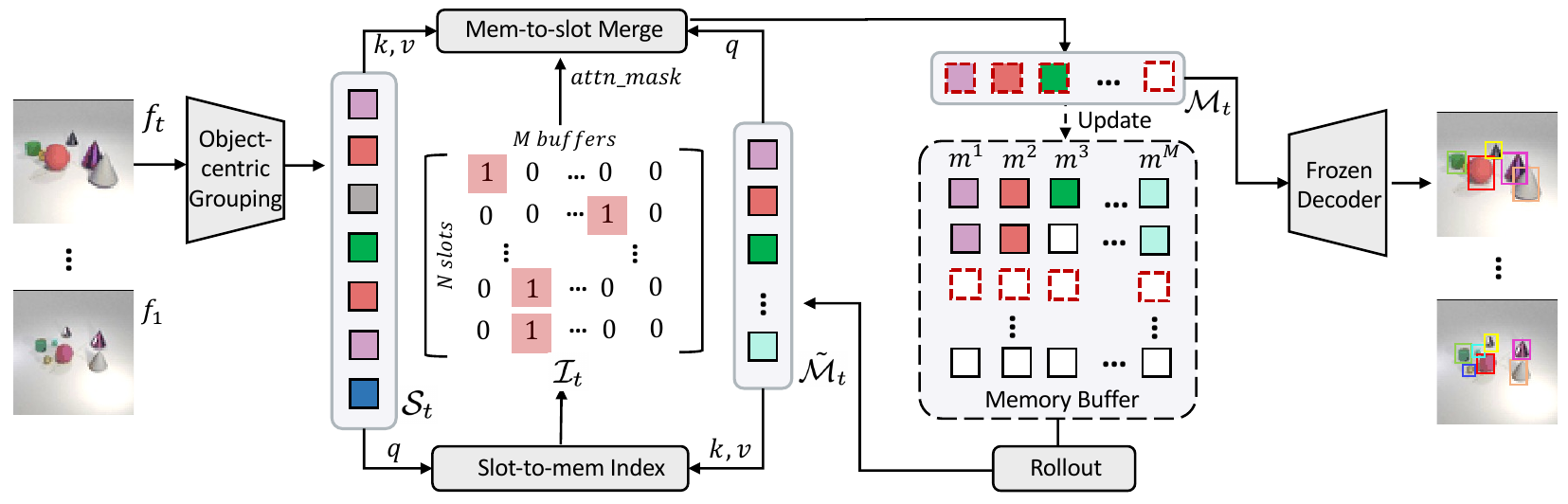}
\caption{\textbf{Overview of OC-MOT}. It consists of two main modules. i) An index-merge module that adapts object-centric slots $\mathcal{S}_t$ into detection results $\mathcal{M}_{t}$ via two steps. First, index each slot into memory buffers by a learnable index matrix $\mathcal{I}_{t}$ indicating all the slot-to-memory assignments. Second, merge slots assigned to the same buffer by recalculating the attention weights masked by $\mathcal{I}_{t}$ backwards. ii) A object memory module that improves temporal consistency by rolling historical state forwards for object association. 
For MOT evaluation, we decode $\mathcal{M}_{t}$ to masks or bounding boxes via a frozen decoder in the object-centric grouping module.}
\label{fig:method}
\end{figure*}

\textbf{Unsupervised Object-centric Learning.} 
Unsupervised object-centric learning describes approaches which aim at tackling the binding problem of visual input signals to objects without additional supervision~\cite{greff2020binding}.
This is often accomplished using architectural inductive biases which force the model to encode input data into a set-structured bottleneck where object representations exert competition~\cite{eslami2016attend, locatello2020object, von2020towards} or exclusive binding to features~\cite{greff2017neural, greff2019multi, burgess2019monet, Engelcke2020GENESIS}.
Since their initial development on synthetic image data, these approaches have been extended to more complicated images by adapting the reconstruction objective~\cite{singh2022illiterate, seitzer2022bridging}, to the decomposition of 3D scenes~\cite{chen2021roots, niemeyer2021giraffe, stelzner2021decomposing}, to synthetic videos~\cite{Kosiorek2018SQAIR,Jiang2020SCALOR, crawford2020exploiting,kabra2021simone,kipf2022conditional,singh2022simple} and to real-world videos by exploiting additional modalities and priors~\cite{kipf2022conditional, bao2022discovering, elsayed2022savi++}.
Our work is most closely related to the last group of methods which apply object-centric learning methods to real-world videos, yet in contrast does not focus on the derivation of object-centric representations themselves.
Instead we focus on how object-centric representations can be used to perform multiple object tracking via long-term memory.
Our work presents the first dedicated memory module, which, independent of the origin of the object-centric representation can match occurrences of objects to previously discovered objects and thus track these over time.

%
%

\looseness=-1\textbf{Self-supervised MOT.} Most works study MOT in supervised settings, where the models are trained with object-level bounding box labels and ID labels~\cite{chu2019famnet, zhang2019robust,zhou2020tracking,zeng2022motr,cai2022memot}. Tracktor++~\cite{bergmann2019tracking} uses a ready-made detector\cite{girshick2015fast} to generate object bounding boxes and propagates them to the next frame as region proposals. MOTR~\cite{zeng2022motr} simultaneously performs object detection and association by autoregressively feeding a set of track queries into a Transformer decoder at the next timestep. To reduce the hand-label annotations, several recent approaches leverage the self-supervised signals to learn object associations from widely available unlabeled videos. For example, CRW~\cite{wang2019learning} and JSTG~\cite{zhao2021modelling} learns video correspondences by applying a cycle-consistent loss. Without  fine-tuning, these models track objects at inference time by propagating the annotations from the first frame.

Our work is mostly related to the unsupervised detect-to-track approaches that assume a robust detector is available. SORT~\cite{bewley2016simple} and IOU~\cite{bochinski2017high} associate detections using heuristic cues such as Kalman filters and intersection-of-union of bounding boxes. Such models do not need training but fail to handle scenarios with frequent occlusion and camera motion. A recent related method uses cross-input consistency~\cite{bastani2021self} to train the tracker: given two distinct inputs from the same video sequence, the model is encouraged to produce consistent tracks. Unfortunately, it suffers performance degradation once the detection boxes are not accurate, e.g., the grouping results from the object-centric model. 
For both supervised and unsupervised trackers, large amount of detection labels are required to train a strong detector. Additionally, supervised trackers require ID labels train feature representations. Overall, MOT is a label-heavy pipeline. Our work has the potential reduce the labeling cost. The heavy-lifting part of object localization and feature binding are done in a self-supervised way: on both backbone training and grouping training.

\looseness=-1\textbf{Memory Models.} Memory models have been widely used in many video analysis and episodic tasks, such as  action recognition~\cite{wu2019long,jin2021temporal}, video object segmentation~\cite{lu2020video,oh2019video,lai2020mast}, video captioning~\cite{pei2019memory}, reinforcement learning~\cite{goyal2022retrieval}, physical reasoning~\cite{alias2021neural}, and code generation~\cite{liu2021retrieval}. These works utilize an external memory to store  features of prolonged sequences, which can be time-indexed for historical information integration. Recently, memory models have also been used to associate objects for video object tracking. MemTrack~\cite{yang2018learning} and SimTrack~\cite{fu2021stmtrack} retrieve useful information from the memory and combine it with the current frame. However,  they ignore the inter-object association and only focus on single object tracking. MeMOT~\cite{cai2022memot} builds a large spatial-temporal memory for MOT, which stores all the identity embeddings of tracked objects and aggregates the historical information as needed. As expected, it requires costly object ID labels for training the memory. In this paper, we propose a self-supervised memory that leverages the memory rollout for object association. In contrast to previous learnable memory modules, our approach does not write global information in the memory via gradient descent~\cite{trauble2022discrete} but rather maintains a running summary of the scene similar to~\cite{Goyal2021RIMs} (but with multi-head attention rollout). Different than~\cite{Goyal2021RIMs}, we explicitly enforce an assignment between objects and memory buffers with subsequent merging steps for MOT. 
\section{Method}

Our OC-MOT improves over traditional OCL frameworks in terms of tracking objects as a whole, and consistently over time. This is achieved by extending the traditional OC framework with a self-supervised memory to: i) Store historical information in the memory to fight against noise and occlusion. This helps improve temporal consistency. ii) Use the complete representation read-out from the memory to consolidate parts captured in different slots, which resolves the part-whole problem. The overall framework of OC-MOT is shown in Figure~\ref{fig:method}. 
Given slots $\{\mathcal{S}_t\}_{t=1}^{T}$ extracted from $T$ video frames by an object-centric grouping module, OC-MOT first uses the memory rollout $\tilde {\mathcal{M}}_t$ to perform slot-to-memory indexing. Then, it merges the slots as $\mathcal{M}_t$ to update the memory. 

\subsection{Object-centric Grouping}
The object-centric grouping module uses Slot Attention\cite{locatello2020object} to turn the set of encoder features from video frames  into a set of slot vectors $\{\mathcal{S}_t\}_{t=1}^{T}$. The model is trained with a self-supervised reconstruction loss $L_{oc\_rec}=||y - \text{Dec}(\mathcal{S})||^2$, where $y$ can be the raw frame pixels, or feature representations extracted from the frames. The decoder has a compete-to-explain inductive bias to encourage binding of objects into individual slots.

\subsection{Memory Module}

We store the historical representations of all tracked objects into memory buffers $\mathcal{M} \in \mathbb{R}^{M\times T \times d}$ where $M$ is the buffer number and $d$ denotes the representation dimension. The memory is implemented with a first-in-first-out data structure and reserves a maximum of $T_{max}$ time steps for each object. At time step $t$, the detection results are $\mathcal{M}_t = \{m_t^1,...,m_t^M \}$ if we denote $m_t$ as the object representation. 
Intuitively, each buffer is a tracklet.

\vspace{+2.5mm}
\noindent \textbf{Memory rollout.} At time step $t$, the memory rolls the past states forward, and predicts the current object representations for all slots to index. The rollout process integrates the multi-view object representations together and handles the part-whole matching in the occlusion scenarios.
Without losing generality, we denote all the past representations as  $\mathcal{M}_{<t}$. The rollout $\tilde {\mathcal{M}}_t \in \mathbb{R}^{M\times d}$ is obtained by:
\begin{equation}
    \tilde {\mathcal{M}}_t = \text{Rollout}(\mathcal{M}_{<t}).
\end{equation}
We adopt a mini GPT-2 model~\cite{radford2019language} containing only 1.6M parameters as the rollout module. It performs temporal reasoning via an auto-regressive transformer.

\subsection{Index-Merge Module}
\begin{figure}[!ht]
	\centering
	\includegraphics[width=80mm]{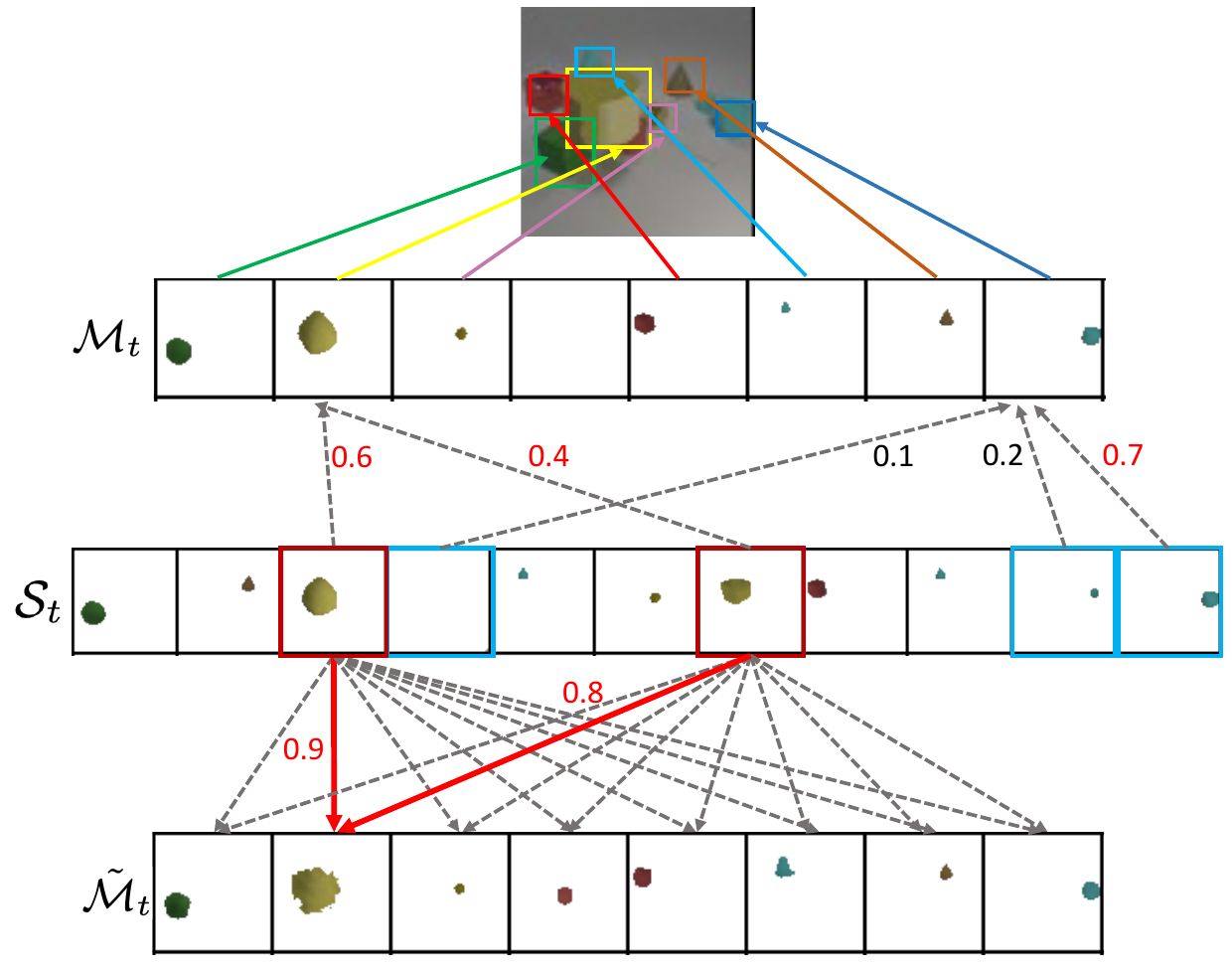}
 \vspace{+0.3cm}
\caption{\textbf{Visualization of the index-merge module}. Index step: we show how to generate an index matrix from slots $\mathcal{S}_t$ to memory rollout $\tilde{\mathcal{M}}_t$. Note that the duplicate slots (red boxes) or similar slots (blue boxes) may be assigned to the same buffer. Merge step: the model recalculates the attention weights for slot merging, masked by the index matrix. The wrongly assigned slots can be filtered out with very low weights.
 }
\label{fig:intro}
\end{figure}

The index-merge module is used as a discrete interface between memory buffers and slots. To achieve this, we split the object association process into the index step and merge step, as shown in Figure~\ref{fig:intro}, which can be achieved by standard multi-head attention (MHA)~\cite{vaswani2017attention} blocks. 
\vspace{-3mm}
\paragraph{Slot-to-memory index.} The index matrix 
$\mathcal{I}_{t} \in \mathbb{R}^{N\times M}$ 
indicates soft slots-to-buffer assignment. 
To compute it, 
we train a MHA
block that takes the slots $\mathcal{S}_t \in \mathbb{R}^{N\times d}$ as query, and rollout $\tilde {\mathcal{M}}_t$
as key and values, where $N$ is slot number:
\begin{equation}
    \mathcal{I}_{t}=\text{MHA}(k,v=\tilde {\mathcal{M}}_t, q=\mathcal{S}_t).\text{attn\_weight}
\end{equation}

\paragraph{Memory-to-slot merge.} 
Our goal is to make sure a buffer represents one object by pooling from the slots that belong to that object, while simultaneously dealing with slots that represent parts of an object or duplicates. Thus,
we stack another MHA
block to merge the slots, using $\mathcal{I}_t$ as masked attention weights.  Specifically, the merging function is defined as below:
\begin{equation}
    m_t=\text{MHA}(k,v=\mathcal{S}_t, q=\tilde {\mathcal{M}}_t, attn\_mask=\mathcal{I}_t).
\end{equation}
Here, the query is the rollout $\tilde {\mathcal{M}}_t$; the key and value are slots $\mathcal{S}_t$. We apply $\mathcal{I}_t$ as the attention mask in MHA such that the re-normalized attention weights can be used for merging. This helps us to deal with wrongly-assigned slots. For example, if there are three slots and two of them are matched to one buffer, the attention weight could be $[0.8, 0.2, 0]$ indicating that the second slot does not belong to this buffer.

\subsection{Model Training under EM Paradigm } 
\noindent \textbf{Losses.} The key of training detect-to-track models is to minimize the assignment costs for object associations. Usually, the weights of the pre-trained detector are frozen during training~\cite{cai2022memot,bergmann2019tracking}. Therefore, in our scenario, we freeze the object-centric model and only train the memory module.
Assume we use $\mathcal{L}_{assign}$ to   measure the assignment costs between  slots $\mathcal{S}_t \in \mathbb{R}^{N\times d}$ and memory buffers $\mathcal{M}_t \in \mathbb{R}^{M\times d}$.  The training loss can be formulated as:
\begin{gather}
\mathcal{L}_{MOT} = \sum_{t=1}^{T}\sum_{i=1}^{N} \mathds{1}[Z_t[i] = j]\mathcal{L}_{assign}(\mathcal{S}_t^i, \mathcal{M}_t^j),
\label{eq:1}
\end{gather}
where $\mathcal{Z}_t\in \mathbb{R}^{N}$ denotes the assignments and $\mathcal{Z}_t[i] = j$ means the $i^{th}$ slots matches to the $j^{th}$ buffer. Specifically, we have three options to calculate the assignment cost: 1) use a binary cross-entropy loss on the decoded masks to promote the consistency of object attributes such as shape; 2) use a pixel-wise squared reconstruction loss on the object reconstructions (pixel reconstructions multiplied by object masks) to learn the color information; 3) use the same loss as 2) but directly apply on the feature space. The assignment cost could be a combination of the three losses:
\begin{equation}
\begin{aligned}
    &\mathcal{L}_{assign}(\mathcal{S}_t^i, \mathcal{M}_t^j) = \lambda_1 BCELoss(\text{Dec}(\mathcal{S}_t^i),\text{Dec}(\mathcal{M}_t^j)) \\
    &+ \lambda_2 || \text{Dec}(\mathcal{S}_t^i)-\text{Dec}(\mathcal{M}_t^j)||^2 + \lambda_3 || \mathcal{S}_t^i-\mathcal{M}_t^j||^2,
\end{aligned}
\end{equation}
where $\lambda_1, \lambda_2, \text{and}\  \lambda_3$ are the balancing weights. We use the frozen decoder from the object-centric model to decode object representations into pixel reconstructions and masks.

\vspace{+2.5mm}
\noindent \textbf{Optimization.}
In contrast to prior supervised trackers~\cite{cai2022memot, zeng2022motr} that use ID labels to find the assignments, our model learns the index matrix $\mathcal{I}_t$ without any supervision. One naive solution is to convert $\mathcal{I}_t\in \mathbb{R}^{N\times M}$  to $\mathcal{Z}_t \in \mathbb{R}^{N}$ by performing argmax along the buffer dimension. However,  the argmax function is non-differentiable. 
Even though we apply the straight-through trick~\cite{jang2016categorical} to make it trainable, the optimization easily gets stuck in a local minimum because the model has no chance to evaluate other possible assignments.
To tackle this problem, we take inspiration from the Expectation-maximization (EM) paradigm which optimizes the assignments from seeing all possible assignments in $\mathcal{I}_t$. We formulate the expectation of $\mathcal{S}_t$ matches to $\mathcal{M}_t$ as:
\begin{gather}
\begin{aligned}
    Q(\theta^*, \theta) &= \mathop{\mathbb{E}} [\ln p(\mathcal{S}_t, \mathcal{M}_t|\theta ^*)] \\
    &= \Sigma_{i}\Sigma_{j} p(\mathcal{M}_t^j | \mathcal{S}_t^i) \ln p(\mathcal{S}_t^i, \mathcal{M}_t^j|\theta ^*) \\
    & = -\Sigma_{i}\Sigma_{j} \mathcal{I}_t[i,j]\mathcal{L}_{assign}(\mathcal{S}_t^i, \mathcal{M}_t^j).
\end{aligned}
\end{gather}
Here, $\theta$ is the learnable parameters in the memory module. $p(\mathcal{M}_t^j | \mathcal{S}_t^i)$ denotes the probability of the $i^{th}$ slot is assigned to the $j^{th}$ buffer, which, in our model, it exactly equals $\mathcal{I}_t[i, j]$. Further, we can use $\mathcal{L}_{assign}$ to represent $\ln p(\mathcal{S}_t^i, \mathcal{M}_t^j|\theta ^*)$.
We optimize the parameters of the model in order to maximize the expectation via SGD~\cite{ruder2016overview}, for which we rewrite equation~\eqref{eq:1} as:
\begin{gather}
\mathcal{L} = \sum_{t=1}^{T}\sum_{i=1}^{N}\sum_{j=1}^{M} \mathcal{I}_t[i,j]\mathcal{L}_{assign}(\mathcal{S}_t^i, \mathcal{M}_t^j).
\label{eq:3}
\end{gather}
The above loss~\eqref{eq:3} is applied to both the merged results $\mathcal{M}_t$ and rollout $\tilde {\mathcal{M}}_t$ with each combination weight set as $1$.

\subsection{Model Inference}
During inference, we binarize the indexing matrix $ \mathcal{I}_{t,hard} \in \{0,1\}^{N\times M}$ to strictly assign one slot to one buffer. 
Specifically, $\mathcal{I}_{t,hard}[i,j] = 1$ only if $j = \text{argmax}(\mathcal{I}_{t})[i]$ for $i \in [1,N]$; otherwise, $\mathcal{I}_{t,hard}[i,j] = 0 $. The discrete index supports the object in-n-out logic by indicating the presence of an object.

 \vspace{+2.5mm}
\noindent \textbf{Object-in logic.} For  the first frame, we filter out duplicate slots before using them to initialize memory buffers. Slots with high mask IoU (bigger than $\tau_{iou}$) to other slots will be discarded. For the next  frames, we activate new buffers for new objects if slots have no substantial IoU with any masks of the memory rollout from the last timestep $\{\tilde {\mathcal{M}}_{t-1}^1,...,\tilde {\mathcal{M}}_{t-1}^k \}$, where $k$ is the active buffer number. Note that, for training, we replace the rollout with slots from the last timestep $\{\mathcal{S}_{t-1}^1,...,\mathcal{S}_{t-1}^N \}$ because the rollout is not reliable at the early training stage.

\vspace{+2.5mm}
\noindent \textbf{Object-out logic.} To re-track an object, we keep the buffer alive for $\tau_{out}$ consecutive frames when the object is occluded or disappears. In other words, if an object disappears for more than $\tau_{out}$ frames, the buffer will be terminated.


\section{Experiments}
\looseness=-1We show that 1) OC-MOT consolidates ``objects"  in memory and greatly improves the temporal consistency of object-centric representations; 2) the gap between object-centric learning and MOT can be narrowed down by involving partial labels to improve the grouping performance; 3) the ablation studies demonstrate the effectiveness and feasibility of each module in the  framework. Finally, we turn to KITTI~\cite{Geiger2012CVPR} to discuss our limitations.

\begin{table*}[!ht]
	\centering
\resizebox{0.95\textwidth}{!}{
\begin{tabular}{lccccccccc}
\toprule[1.5pt]
Method  & Detection Label & ID Label  & IDF1 $\uparrow$   & MOTA $\uparrow$     & MT $\uparrow$    & ML $\downarrow$   & FP   $\downarrow$   & FN $\downarrow$    & IDS $\downarrow$  \\\hline
\multicolumn{10}{c}{CATER~\cite{girdhar2019cater}} \\ \hline \hline
SAVi~\cite{kipf2022conditional}   &  &    & 73.2\%  & 52.5\%   & 75.2\% & 21.2\%  &  305027 &  130810 &20352  \\        
IOU~\cite{bochinski2017high} &  &  &  83.0\% &  77.4\%  & 73.3\%  & 17.4\%  &  \textbf{35480} & 173595  & 8259  \\
SORT~\cite{bewley2016simple}   &   &   & 84.5\% & 79.2\%  & 71.8\% & 24.1\%  & 43097 & 148068 & 8219    \\     
Visual-Spatial~\cite{bastani2021self}  & & &  85.8\%     &  80.3\%      &   76.6\%     &     20.8\%      & 51348      &    129680  &     7562       \\
\textbf{OC-MOT}   &  &   & \textbf{88.6\%} & \textbf{82.4\%} &\textbf{82.3\%}       & \textbf{13.9\%} & 57792    & \textbf{105054}   &  \textbf{5658}  \\\hdashline
MOTR~\cite{zeng2022motr}   & 100\%   & 100\%    & 89.3\% & 83.3\%   &  84.8\%      &  4.9\%     & 60647  & 96746  & 3366   \\\hline
\multicolumn{10}{c}{FISHBOWL~\cite{tangemann2021unsupervised} } \\ \hline \hline
SAVi~\cite{kipf2022conditional} &6.25\%    &   & 46.9\% & 32.3\%  & 47.3\% & 15.1\%  & 122006 & \textbf{96710} & 12504 \\  
SORT~\cite{bewley2016simple}   &6.25\% & &68.4\% & 64.3\% & 42.6\%  & 31.9\%  & 30912 & 132434 & 15278  \\  
IOU~\cite{bochinski2017high} & 6.25\%  &  &  71.3\% &  66.6\%  & 11.0\%  & 62.7\%  &31672   &  135394 & 10306   \\
Visual-Spatial~\cite{bastani2021self}  &6.25\%   & &74.6\%    &  68.1\%      &  48.2\%       & 19.8\%   & 28845 & 131076 & 8754    \\
\textbf{OC-MOT} &6.25\%   & &  \textbf{77.9\%} & \textbf{70.3\%} &    \textbf{50.2\%}  &  \textbf{ 13.2\%}  & \textbf{14738}  & 136852 & \textbf{5898}  \\\hdashline
MOTR~\cite{zeng2022motr}   & 100\%   &  100\%  & 81.6\% & 79.8\%   &  58.3\%      &  10.1\%  & 9678   &  92862  & 4185   \\
\bottomrule[1.5pt]
\end{tabular}
}
\vspace{+0.05in}
\caption{\textbf{Evaluation results on CATER and FISHBOWL}. For CATER,  the object-centric grouping module is pre-trained without any label. For FISHBOWL, the grouping module is pre-trained with $6.25\%$ mask labels to improve the detection accuracy. The supervised MOTR~\cite{zeng2022motr} is trained with 100\% box labels and ID labels.
The best results of unsupervised trackers are marked in bold.}
	\label{tab:results}
\end{table*}

\vspace{+2.5mm}
\noindent\textbf{Datasets.} CATER~\cite{girdhar2019cater} is a widely used synthetic video dataset for object-centric learning. It is rendered using a  library of 3D objects with various movements. Tracking multiple objects requires temporal reasoning about the long-term occlusions, a common issue in this dataset. FISHBOWL~\cite{tangemann2021unsupervised} consists of 20,000 training and 1,000 validation and test videos recorded from a publicly available WebGL demo of an aquarium, each with a resolution of 480×320px and 128 frames. Compared to CATER, FISHBOWL records more complicated scenes and has even more severe object occlusions. Besides, we also work on the real-world driving dataset KITTI~\cite{Geiger2012CVPR} to analyze the limitation of the proposed object-centric framework.

\vspace{+2.5mm}
\noindent\textbf{Metrics.} Following the standard MOT evaluation protocols~\cite{ristani2016performance, milan2016mot16}, we use Identity F1 score (IDF1), Multiple-Object Tracking Accuracy (MOTA), Mostly Tracked (MT), Mostly Lost (ML), and Identity Switches (IDS) as the metrics. Specifically, IDF1 highlights the tracking consistency, and MOTA measures the  object coverage. To weight down the effect of detection accuracy and focus on the association performance, we set the IoU distance threshold as 0.7.
We also introduce Track mAP~\cite{dave2020tao}, which is more sensitive to identity switches by matching the object bounding boxes to ground-truth through the entire video using 3D IoU. 
 
\begin{figure}[!t]
    \centering	\includegraphics[width=62mm]{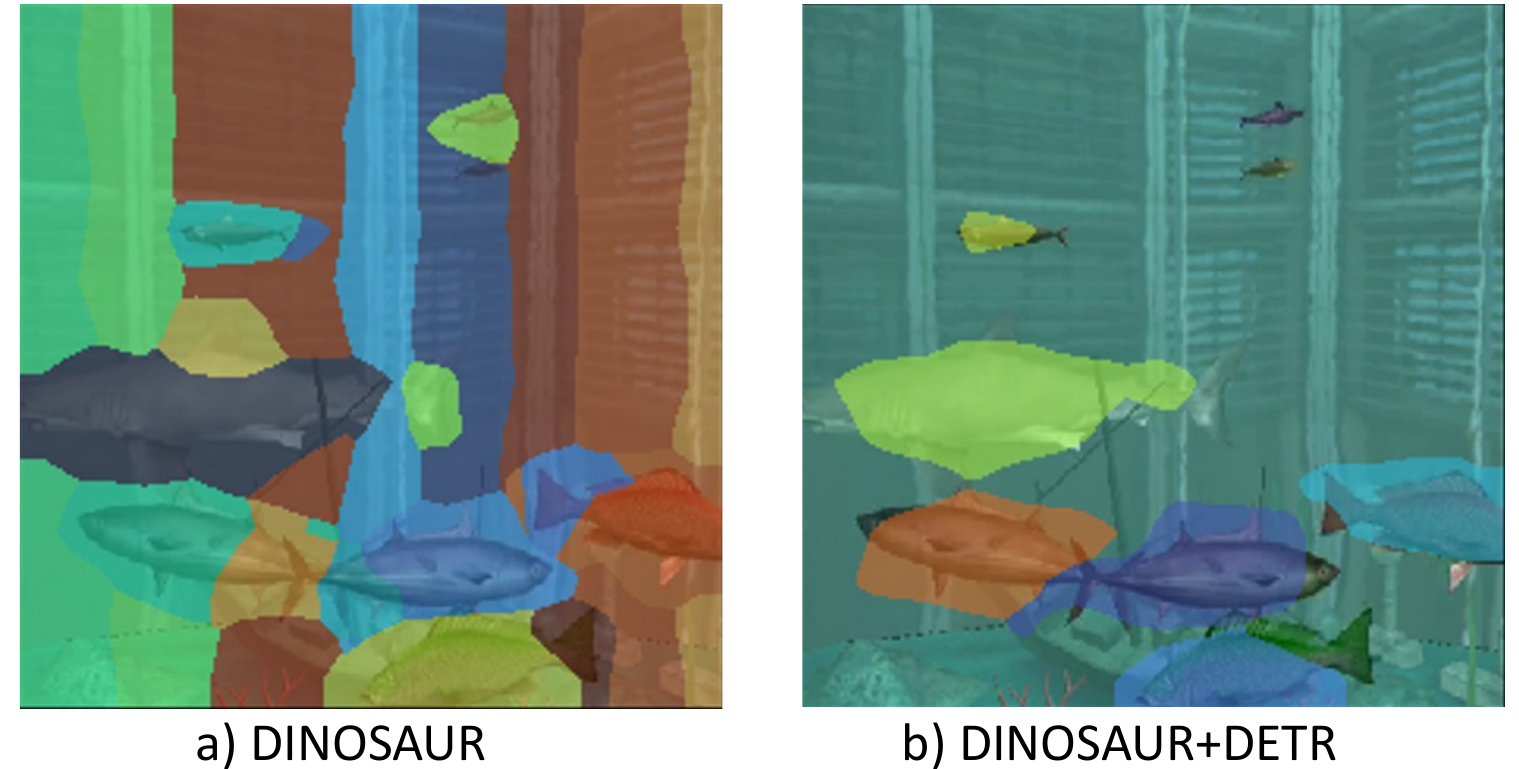}
\caption{Comparisons of different grouping module settings on FISHBOWL: a) self-supervised DINOSAUR has high object recall but over-segments on both background and big fishes, causing trouble to memory buffer initialization. b) Tuning DINOSAUR with supervised DETR loss and partial mask labels resolves the over-segmentation issue and filters out background slots. 
 }
\label{fig:group}
\end{figure}

\vspace{+2.5mm}
\noindent\textbf{Implementation details.} We train OC-MOT using the Adam optimizer~\cite{kingma2014adam} with a learning rate of $2\cdot10^{-4}$ and an exponentially decaying learning rate schedule. The models were trained on 8 NVIDIA GPUs with batchsize of 8. We set $\tau_{out}$ as 5 for buffer termination.
The IoU threshold $\tau_{iou}$ is set as $0.9$. For the experiments on CATER, we pretrain a SAVi model for object grouping without any annotation. We set $N=11$ and $M=15$.
The hyperparameters in the training loss  $\lambda_1, \lambda_2, \lambda_3$ are selected as $1$, $0.1$, $0$. For the experiments on FISHBOWL, we used a pretrained image-level DINOSAUR~\cite{seitzer2022bridging} as the grouping module and selected  $\lambda_1, \lambda_2, \lambda_3$ as $1$, $0$, $1$. We set $\lambda_2$ to $0$ due to GPU memory limitations when combining the EM loss computation with the high dimensional DINOSAUR features. We set $N= 24$ and $M=40$. In complex scenes of FISHBOWL, we noticed a performance drop due to more severe part-whole issues and over-segmentation on the background as illustrated in Figure~\ref{fig:group}. To avoid tracking background objects and reduce over-segmentation on big objects, we suggest further improving object-centric grouping by utilizing temporal sparse labels. To be more specific, we apply supervised DETR~\cite{carion2020end}-style loss on the decoded masks of slots. Since the object grouping loss already takes the heavy-lifting of discovering objects and parts, we only require very few mask labels to inject semantics about which objects are interesting and how to merge parts into a whole object. In practice, we utilized $6.25\,\%$ (randomly label 8 frames in 128-frame videos) mask labels for DINOSAUR pre-training, with both DETR loss and self-supervised reconstruction loss.

\begin{table}[!ht]
\centering
\resizebox{80mm}{!}{
\begin{tabular}{lcccc}
\toprule[1.5pt]
\multirow{2}{*}{Method} & OC Metric & \multicolumn{3}{c}{MOT Metric} \\ \cmidrule(r){2-2} \cmidrule(r){3-5}
                        & FG-ARI $\uparrow$    & IDF1 $\uparrow$    & MOTA $\uparrow$   & Track mAP $\uparrow$ \\
                        \cmidrule(r){1-1}\cmidrule(r){2-2} \cmidrule(r){3-5}
SAVi                    & 90.2      & 72.3    & 52.5   & 42.8        \\
OC-MOT                  & \textbf{93.8}      & \textbf{88.6}    & \textbf{82.4}   & \textbf{66.2}        \\ \bottomrule[1.5pt]
\end{tabular}
}
\vspace{+0.05in}
\caption{Comparisons with video object-centric models on CATER. Note that FG-ARI~\cite{kipf2022conditional} is a commonly used OC metric.}
	\label{tab:results_cater}
\end{table}

\subsection{Comparison with the State-of-the-art Methods}
\noindent\textbf{Baselines.} We compare OC-MOT with one object-centric method (SAVi~\cite{kipf2022conditional}), three unsupervised MOT methods (IOU~\cite{bochinski2017high}, SORT~\cite{bewley2016simple}, and Visual-Spatial~\cite{bastani2021self}), and one fully supervised MOT method (MOTR~\cite{zeng2022motr}). For the SAVi evaluation, we remove the background slots and treat each slot prediction as a tracklet. When training SAVi on FISHBOWL, we also provide 6.25\% temporal sparse mask labels to be comparable to our own setting. For fair comparisons, we use the same pre-trained object-centric model (DINOSAUR with 6.25\% detection labels) as the detector for IOU, SORT, and Visual-Spatial.
MOTR utilizes a query-based transformer for both object detection and association but requires object-level annotations (both boxes and object ID) for model training. 
This model and its follow-ups have achieved SOTA results on several MOT benchmarks.

\begin{figure*}[!t]
	\centering
	\includegraphics[width=0.9\textwidth]{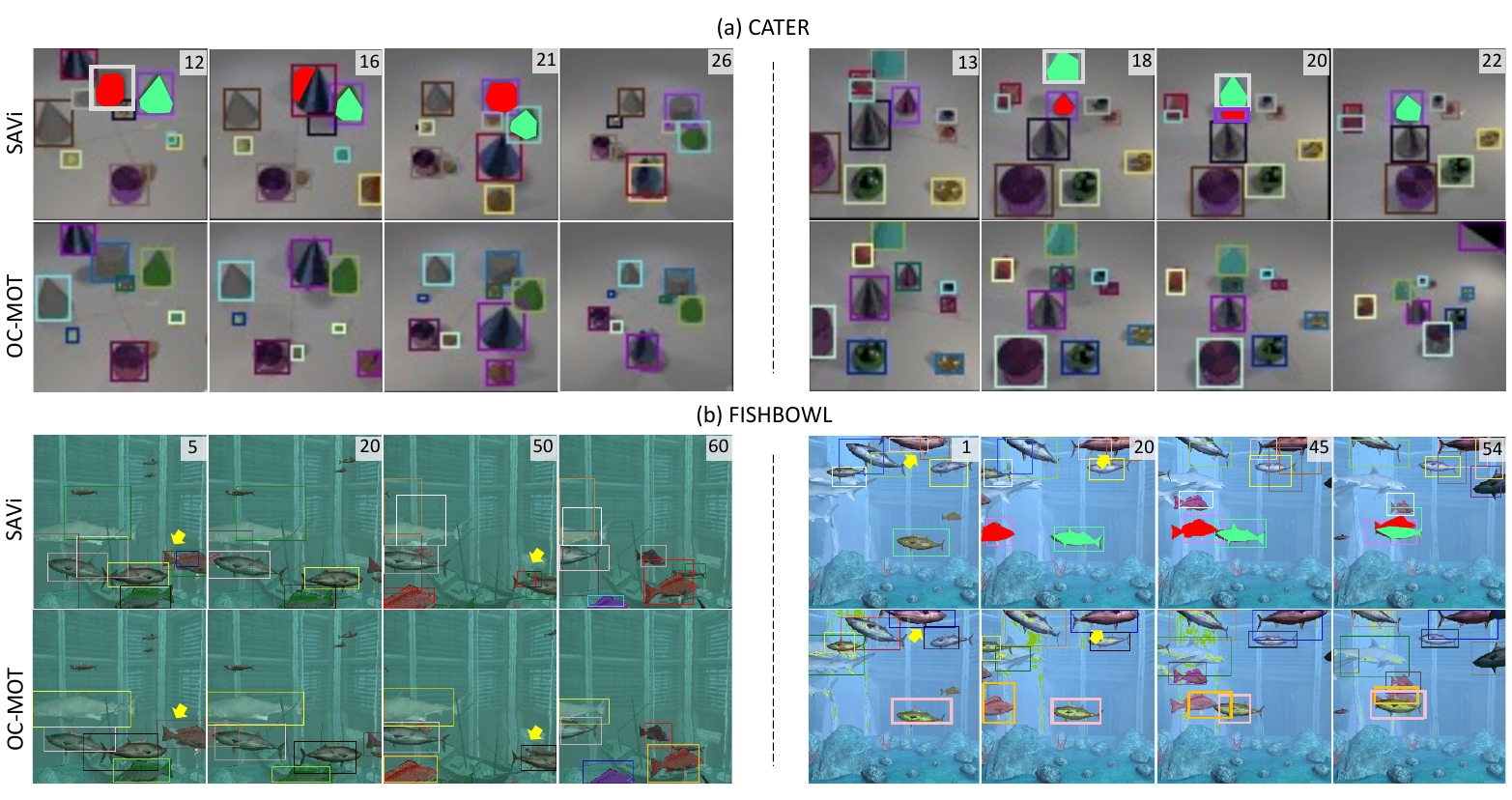}
\caption{\textbf{MOT results on CATER and FISHBOWL.} We highlight the occlusion cases with colored masks. SAVi over-segments the objects (yellow arrows) and has ID switches after occlusions. In contrast, OC-MOT tracks objects more consistently over time.
 }
\label{fig:vis}
\end{figure*}

\vspace{+2.5mm}
\noindent\textbf{Results on CATER.}
As shown in Table \ref{tab:results}, OC-MOT substantially outperforms the video object-centric model and other unsupervised baselines. Our approach is also competitive with supervised MOTR~\cite{zeng2022motr} trained on expensive object-level annotations, yielding only slightly lower IDF1 and MOTA. OC-MOT can keep tracking more objects but produces fewer ID switches. For example, it achieves $82.3\,\%$ Mostly Tracked (MT) and $13.9\,\%$ Mostly Lost (ML), and shows only $5658$ IDS. Moreover, in Table~\ref{tab:results_cater}, SAVi achieves 90.2\% of FG-ARI but performs bad in terms of other MOT metrics such as 42.8\% Track mAP, indicating that the FG-ARI is not a good metric for measuring object-level temporal consistency.

\vspace{+2.5mm}
\noindent\textbf{Results on FISHBOWL.} FISHBOWL is a more challenging benchmark with serious occlusions and complicated backgrounds. 
In this scenario, SAVi~\cite{kipf2022conditional} tends to split the complicated background into multiple slots, causing the high number of false positive (FP).
Table~\ref{tab:results} shows that OC-MOT achieves state-of-the-art performance among the unsupervised tracking methods. By getting a much lower IDS number, our approach shows its advantage in solving the occlusion problem. The non-linear transformation in frequent occlusions cannot be handled by IoU-based association (IOU~\cite{bochinski2017high}) or Kalman filter (SORT~\cite{bewley2016simple}). 
Compared to supervised MOTR, what would like to highlight is the impressive association capability of OC-MOT. We point out that the lower IDF1 and MOTA are mainly caused by the detection limitation of existing OC models (e.g., DINOSAUR decoder predicts masks at a low feature resolution, making it hard to get pixel-level accuracy).

\begin{figure}[!h]
	\centering
	\includegraphics[width=82mm]{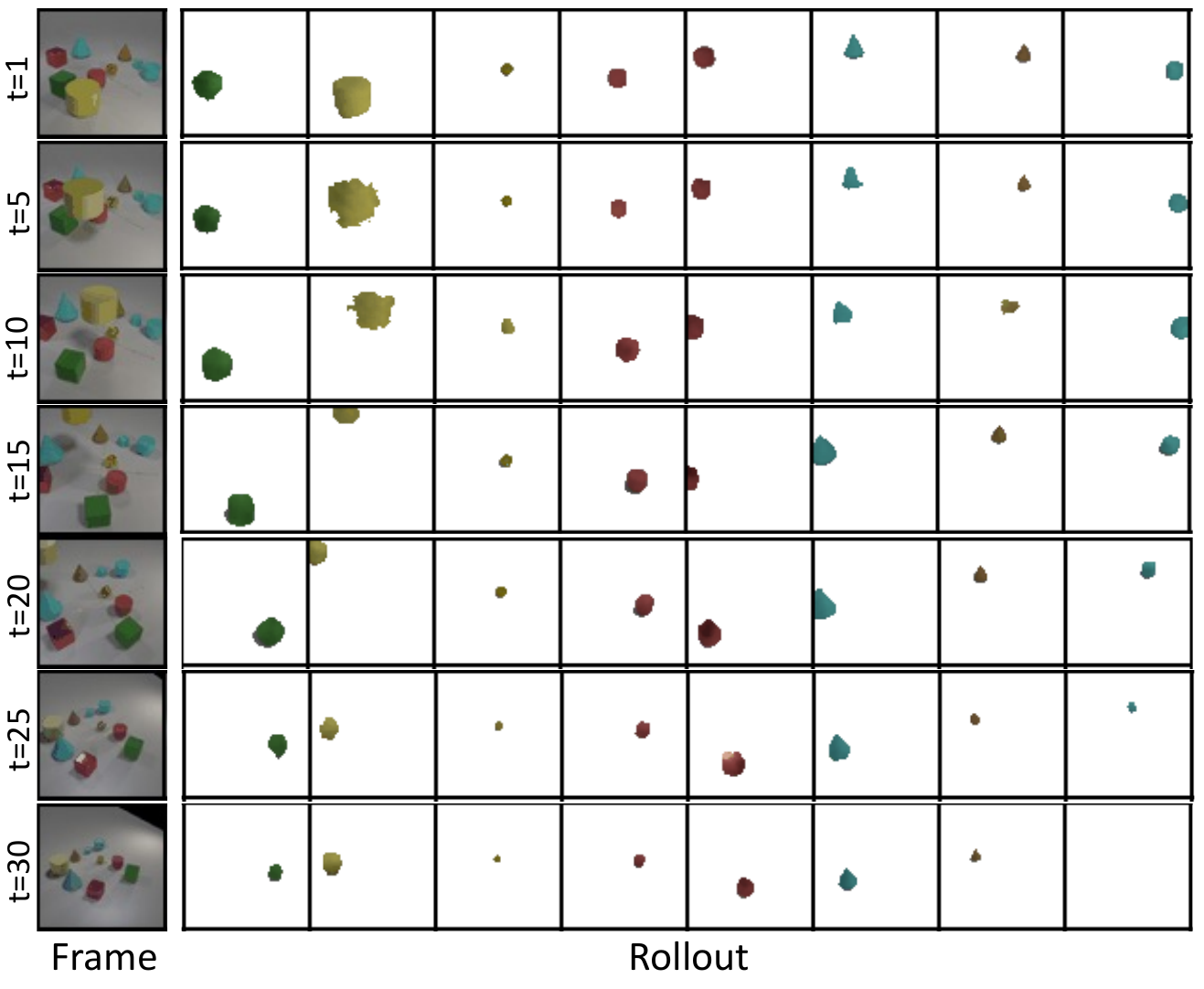}
\caption{\textbf{Visualization of memory rollout.} We show the object reconstructions decoded from the rollout representations. Each column denotes a memory buffer. The rollout predictions are consistent and complete, even when objects are partially occluded
 }
\label{fig:rollout}
\end{figure}

\subsection{Visualization}
The MOT results on the occlusion cases are visualized in 
Figure~\ref{fig:vis}. OC-MOT associates the slots from the object-centric model and
generates consistent predictions even when objects frequently interact with each other. Due to the severe occlusions, SAVi~\cite{kipf2022conditional} fails to track objects even using the track query as input, thereby causing more ID switches. Moreover, SAVi produces more false positives due to over-segmentation.

In Figure~\ref{fig:rollout}, we visualize the memory rollout results by decoding the representations to object reconstructions. The memory starts to roll out after the first frame, and, at $t=1$, we visualize the existing memory features. We can observe that the rollouts achieve good temporal consistency and, even more interestingly, that the memory can predict a  complete object even when it has been partially occluded.

\subsection{Ablation Studies}
\label{sec:abl_oc}

\noindent\textbf{Component analysis.}
Table~\ref{tab:component} compares different design choices for the key components in OC-MOT. For the index-merge module, a naive solution is to use a parameter-free dot-product to calculate the feature similarity, inspired by~\cite{rahaman2021dynamic}. As expected, it produces the worst association performance. A further option is to train one single MHA (i.e., two MHAs with shared weights) to cluster slots to buffers as in~\cite{Goyal2021RIMs}. To get the discrete index for object in and out logic, we still follow the indexing and merging steps yet only calculating the attention weights once. We observe that this model yields slightly lower IDF1 and MOTA than training two MHA modules. The latter choice is mathematically the similar but with higher module capacity. For the memory module, we compare utilizing the rollout module to only using the last tracks as the index query. Without aggregating the historical memory features, the association performance drops dramatically, indicating the necessity of building a memory to handle the MOT problem.

\begin{table}[!ht]
\centering
\resizebox{80mm}{!}{
\begin{tabular}{cc|ccc}
\toprule[1.5pt]
 \begin{tabular}[c]{@{}c@{}}Index-Merge\\ Module\end{tabular}    & \begin{tabular}[c]{@{}c@{}}Memory\\ Module\end{tabular} & IDF1 $\uparrow$  & MOTA $\uparrow$  & IDS $\downarrow$  \\ \hline
Dot-product & Rollout  & 72.0\% & 61.5\% & 22050 \\
 One MHA & Rollout       &  86.2\%    &80.5\%   &7655 \\
Two MHA & Last Tracks       &  77.2\%    & 68.8\%     & 16582\\\hdashline
Two MHA         & Rollout   & \textbf{88.6\%} & \textbf{82.4\%} & \textbf{5658}  \\ \bottomrule[1.5pt]
\end{tabular}
}
\vspace{+0.3cm}
\caption{Ablation on OC-MOT components on CATER.}
\label{tab:component}
\end{table}

\noindent\textbf{Effect of memory length.} In Table~\ref{tab:length}, we explore the effect of the memory length $T_{max}$ from 6 to 32. Note that $T_{max}$ equals the length of the training sequence. The tracking performance increases as $T_{max}$ grows. However, for longer videos, we should set a max length of memory due to hardware limitations. To make the model more applicable, we propose to reserve a short-term memory trained with sequences sampled by slow-to-fast pace. The various sampling rates produce both short-term and long-term information and, more importantly, include the occlusion cases during training. Quantitatively, this sampling strategy peaks in performance with $T_{max}=6$. Unless noted otherwise, we set $T_{max}$ to 6 as default in other experiments.

\begin{table}[!ht]
\centering
\resizebox{80mm}{!}{
\begin{tabular}{cc|ccc}
\toprule[1.5pt]
$T_{max}$  & Sequence Sampling                       & IDF1 $\uparrow$  & MOTA $\uparrow$  & IDS  $\downarrow$  \\ \hline
6  & Consecutive                    & 82.9\% & 76.3\% & 7601 \\
10  & Consecutive                    & 83.2\% & 76.5\% & 7524 \\
20  & Consecutive                    & 86.9\% & 78.1\% & 6230 \\
32 & Consecutive                    & 88.4\% & 82.1\%     &  5763      \\\hdashline
6  & Slow-Fast & \textbf{88.6\%} & \textbf{82.4\%} & \textbf{5658}   \\ \bottomrule[1.5pt]
\end{tabular}
}
\vspace{+0.3cm}
\caption{Ablation on the memory length on CATER.}
\label{tab:length}
\end{table}

\subsection{Limitations}

There exist some limitations of the proposed OC-MOT.

\begin{figure}[!h]
	\centering
	\includegraphics[width=83mm]{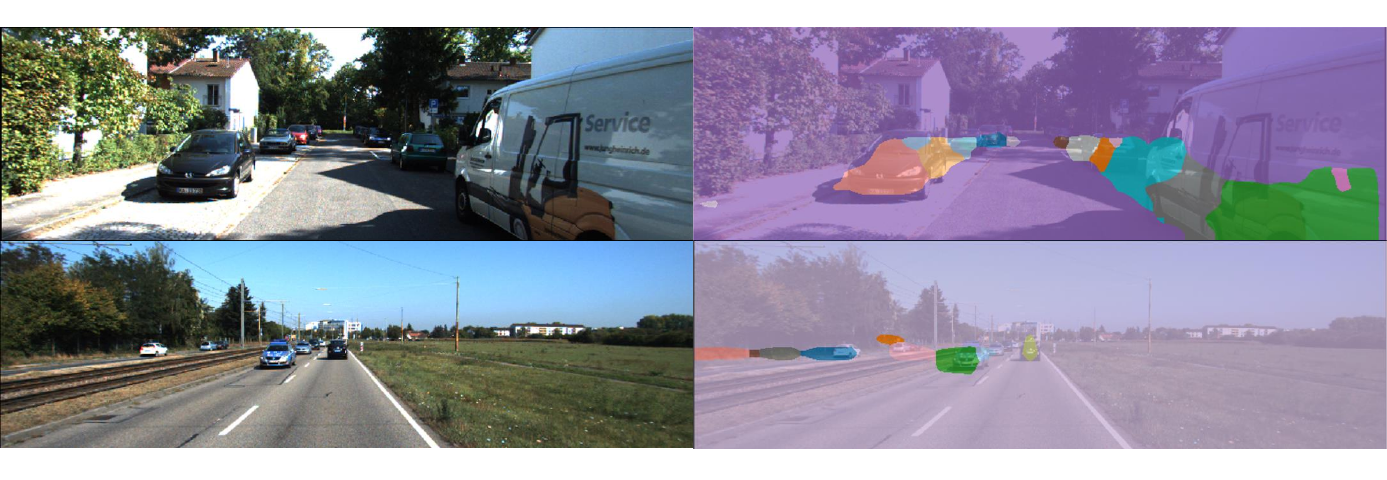}
\caption{DINOSAUR grouping results on KIITI. We observe that the masks are imprecise, espcially for objects that are far away.
 }
 \vspace{-2mm}
\label{fig:kitti}
\end{figure}

\vspace{+1mm}
\noindent\textbf{Inductive bias in the grouping architecture.} We apply the same DINOSAUR+DETR grouping module with 6.25\% temporally sparse mask labels to KITTI dataset. Figure \ref{fig:kitti} visualizes the grouping results. The cars can be detected but the predicted masks are not accurate, especially for far-away objects.
One reason is that DINOSAUR predicts masks at a feature resolution that is down-scaled 16 times from the original size. The architecture of the grouping module needs to be further improved considering multi-resolution inductive biases that have already been adopted in  supervised detection and segmentation pipelines. 
We encourage researchers to develop stronger OC models with powerful detection performance but low labeling cost.

\vspace{+2.5mm}
\noindent\textbf{The model is not trained end-to-end.}
In this paper, we use the pre-trained OC model as a plug-n-play detector, which is supposed to handle different data flexibly. Potential future work is to extend OC-MOT into an end-to-end framework. The object prototype built in the memory may be useful as a prior for object discovery.
\section{Conclusion}
\looseness=-1In this paper, we build a pipeline for MOT with object-centric backbones. With memory modules, we can address both part-whole issues and consistently track objects over time. Overall, our approach improves over conventional tracking-by-detection pipelines by replacing expensive annotations (especially ID annotations) with self supervision. This work opens many directions for new research. First of all, it allows for active learning. For example, the model could elicit a request for labeling on specific frames, further reducing necessity for costly annotations.
Furthermore, incorporating memory information as top-down reasoning prior for the object-centric encoder still remains to be explored.
Additionally, we still require few masks and class labels to resolve over-segmentation. Those semantic signals could be distilled from multi-modal foundation models trained with weaker supervision signals (e.g., captioned images). Finally, our results delineate a clear benefit in improving (video) object-centric backbones. 
As we have demonstrated, improvements in self-supervised object-centric learning can greatly facilitate complex downstream vision tasks like MOT, improving performance by training on unsupervised or weakly-supervised data.

{\small
\bibliographystyle{ieee_fullname}
\bibliography{egbib}

\begin{thebibliography}{10}\itemsep=-1pt

\bibitem{bao2022discovering}
Zhipeng Bao, Pavel Tokmakov, Allan Jabri, Yu-Xiong Wang, Adrien Gaidon, and
  Martial Hebert.
\newblock Discovering objects that can move.
\newblock In {\em Proceedings of the IEEE/CVF Conference on Computer Vision and
  Pattern Recognition}, pages 11789--11798, 2022.

\bibitem{bastani2021self}
Favyen Bastani, Songtao He, and Samuel Madden.
\newblock Self-supervised multi-object tracking with cross-input consistency.
\newblock {\em Advances in Neural Information Processing Systems},
  34:13695--13706, 2021.

\bibitem{bergmann2019tracking}
Philipp Bergmann, Tim Meinhardt, and Laura Leal-Taixe.
\newblock Tracking without bells and whistles.
\newblock In {\em Proceedings of the IEEE/CVF International Conference on
  Computer Vision}, pages 941--951, 2019.

\bibitem{bewley2016simple}
Alex Bewley, Zongyuan Ge, Lionel Ott, Fabio Ramos, and Ben Upcroft.
\newblock Simple online and realtime tracking.
\newblock In {\em 2016 IEEE international conference on image processing
  (ICIP)}, pages 3464--3468. IEEE, 2016.

\bibitem{bochinski2017high}
Erik Bochinski, Volker Eiselein, and Thomas Sikora.
\newblock High-speed tracking-by-detection without using image information.
\newblock In {\em 2017 14th IEEE international conference on advanced video and
  signal based surveillance (AVSS)}, pages 1--6. IEEE, 2017.

\bibitem{burgess2019monet}
Christopher~P Burgess, Loic Matthey, Nicholas Watters, Rishabh Kabra, Irina
  Higgins, Matt Botvinick, and Alexander Lerchner.
\newblock {Monet: Unsupervised scene decomposition and representation}.
\newblock {\em arXiv preprint arXiv:1901.11390}, 2019.

\bibitem{cai2022memot}
Jiarui Cai, Mingze Xu, Wei Li, Yuanjun Xiong, Wei Xia, Zhuowen Tu, and Stefano
  Soatto.
\newblock Memot: multi-object tracking with memory.
\newblock In {\em Proceedings of the IEEE/CVF Conference on Computer Vision and
  Pattern Recognition}, pages 8090--8100, 2022.

\bibitem{carion2020end}
Nicolas Carion, Francisco Massa, Gabriel Synnaeve, Nicolas Usunier, Alexander
  Kirillov, and Sergey Zagoruyko.
\newblock End-to-end object detection with transformers.
\newblock In {\em Computer Vision--ECCV 2020: 16th European Conference,
  Glasgow, UK, August 23--28, 2020, Proceedings, Part I 16}, pages 213--229.
  Springer, 2020.

\bibitem{chen2021roots}
Chang Chen, Fei Deng, and Sungjin Ahn.
\newblock Roots: Object-centric representation and rendering of 3d scenes.
\newblock {\em The Journal of Machine Learning Research}, 22(1):11770--11805,
  2021.

\bibitem{chu2019famnet}
Peng Chu and Haibin Ling.
\newblock Famnet: Joint learning of feature, affinity and multi-dimensional
  assignment for online multiple object tracking.
\newblock In {\em Proceedings of the IEEE/CVF International Conference on
  Computer Vision}, pages 6172--6181, 2019.

\bibitem{crawford2020exploiting}
Eric Crawford and Joelle Pineau.
\newblock Exploiting spatial invariance for scalable unsupervised object
  tracking.
\newblock In {\em Proceedings of the AAAI Conference on Artificial
  Intelligence}, volume~34, pages 3684--3692, 2020.

\bibitem{dave2020tao}
Achal Dave, Tarasha Khurana, Pavel Tokmakov, Cordelia Schmid, and Deva Ramanan.
\newblock Tao: A large-scale benchmark for tracking any object.
\newblock In {\em Computer Vision--ECCV 2020: 16th European Conference,
  Glasgow, UK, August 23--28, 2020, Proceedings, Part V 16}, pages 436--454.
  Springer, 2020.

\bibitem{dittadi2022generalization}
Andrea Dittadi, Samuele~S Papa, Michele De~Vita, Bernhard Sch{\"o}lkopf, Ole
  Winther, and Francesco Locatello.
\newblock Generalization and robustness implications in object-centric
  learning.
\newblock In {\em International Conference on Machine Learning}, pages
  5221--5285. PMLR, 2022.

\bibitem{elsayed2022savi++}
Gamaleldin~F Elsayed, Aravindh Mahendran, Sjoerd van Steenkiste, Klaus Greff,
  Michael~C Mozer, and Thomas Kipf.
\newblock Savi++: Towards end-to-end object-centric learning from real-world
  videos.
\newblock {\em arXiv preprint arXiv:2206.07764}, 2022.

\bibitem{Engelcke2020GENESIS}
Martin Engelcke, Adam~R. Kosiorek, Oiwi~Parker Jones, and Ingmar Posner.
\newblock {GENESIS: Generative Scene Inference and Sampling with Object-Centric
  Latent Representations}.
\newblock In {\em International Conference on Learning Representations}, 2020.

\bibitem{eslami2016attend}
SM Eslami, Nicolas Heess, Theophane Weber, Yuval Tassa, David Szepesvari, Koray
  Kavukcuoglu, and Geoffrey~E Hinton.
\newblock {Attend, infer, repeat: Fast scene understanding with generative
  models}.
\newblock {\em Advances in Neural Information Processing Systems}, 2016.

\bibitem{fu2021stmtrack}
Zhihong Fu, Qingjie Liu, Zehua Fu, and Yunhong Wang.
\newblock Stmtrack: Template-free visual tracking with space-time memory
  networks.
\newblock In {\em Proceedings of the IEEE/CVF Conference on Computer Vision and
  Pattern Recognition}, pages 13774--13783, 2021.

\bibitem{Geiger2012CVPR}
Andreas Geiger, Philip Lenz, and Raquel Urtasun.
\newblock Are we ready for autonomous driving? the kitti vision benchmark
  suite.
\newblock In {\em Conference on Computer Vision and Pattern Recognition
  (CVPR)}, 2012.

\bibitem{girdhar2019cater}
Rohit Girdhar and Deva Ramanan.
\newblock Cater: A diagnostic dataset for compositional actions and temporal
  reasoning.
\newblock {\em arXiv preprint arXiv:1910.04744}, 2019.

\bibitem{girshick2015fast}
Ross Girshick.
\newblock Fast r-cnn.
\newblock In {\em Proceedings of the IEEE international conference on computer
  vision}, pages 1440--1448, 2015.

\bibitem{alias2021neural}
Anirudh Goyal, Aniket Didolkar, Nan~Rosemary Ke, Charles Blundell, Philippe
  Beaudoin, Nicolas Heess, Michael~C Mozer, and Yoshua Bengio.
\newblock Neural production systems.
\newblock {\em Advances in Neural Information Processing Systems},
  34:25673--25687, 2021.

\bibitem{goyal2022retrieval}
Anirudh Goyal, Abram Friesen, Andrea Banino, Theophane Weber, Nan~Rosemary Ke,
  Adria~Puigdomenech Badia, Arthur Guez, Mehdi Mirza, Peter~C Humphreys, Ksenia
  Konyushova, et~al.
\newblock Retrieval-augmented reinforcement learning.
\newblock In {\em International Conference on Machine Learning}, pages
  7740--7765. PMLR, 2022.

\bibitem{Goyal2021RIMs}
Anirudh Goyal, Alex Lamb, Jordan Hoffmann, Shagun Sodhani, Sergey Levine,
  Yoshua Bengio, and Bernhard Sch{\"o}lkopf.
\newblock {Recurrent Independent Mechanisms}.
\newblock In {\em International Conference on Learning Representations}, 2021.

\bibitem{greff2019multi}
Klaus Greff, Rapha{\"e}l~Lopez Kaufman, Rishabh Kabra, Nick Watters,
  Christopher Burgess, Daniel Zoran, Loic Matthey, Matthew Botvinick, and
  Alexander Lerchner.
\newblock {Multi-object representation learning with iterative variational
  inference}.
\newblock In {\em International Conference on Machine Learning}, 2019.

\bibitem{greff2017neural}
Klaus Greff, Sjoerd Van~Steenkiste, and J{\"u}rgen Schmidhuber.
\newblock Neural expectation maximization.
\newblock {\em Advances in Neural Information Processing Systems}, 30, 2017.

\bibitem{greff2020binding}
Klaus Greff, Sjoerd Van~Steenkiste, and J{\"u}rgen Schmidhuber.
\newblock {On the binding problem in artificial neural networks}.
\newblock {\em arXiv preprint arXiv:2012.05208}, 2020.

\bibitem{jang2016categorical}
Eric Jang, Shixiang Gu, and Ben Poole.
\newblock Categorical reparameterization with gumbel-softmax.
\newblock {\em arXiv preprint arXiv:1611.01144}, 2016.

\bibitem{Jiang2020SCALOR}
Jindong Jiang, Sepehr Janghorbani, Gerard {de Melo}, and Sungjin Ahn.
\newblock {SCALOR: Generative World Models with Scalable Object
  Representations}.
\newblock In {\em Proceedings of ICLR 2020}, 2020.

\bibitem{jin2021temporal}
Yueming Jin, Yonghao Long, Cheng Chen, Zixu Zhao, Qi Dou, and Pheng-Ann Heng.
\newblock Temporal memory relation network for workflow recognition from
  surgical video.
\newblock {\em IEEE Transactions on Medical Imaging}, 40(7):1911--1923, 2021.

\bibitem{kabra2021simone}
Rishabh Kabra, Daniel Zoran, Goker Erdogan, Loic Matthey, Antonia Creswell,
  Matt Botvinick, Alexander Lerchner, and Chris Burgess.
\newblock Simone: View-invariant, temporally-abstracted object representations
  via unsupervised video decomposition.
\newblock {\em Advances in Neural Information Processing Systems},
  34:20146--20159, 2021.

\bibitem{kingma2014adam}
Diederik~P Kingma and Jimmy Ba.
\newblock Adam: A method for stochastic optimization.
\newblock {\em arXiv preprint arXiv:1412.6980}, 2014.

\bibitem{kipf2022conditional}
Thomas Kipf, Gamaleldin~Fathy Elsayed, Aravindh Mahendran, Austin Stone, Sara
  Sabour, Georg Heigold, Rico Jonschkowski, Alexey Dosovitskiy, and Klaus
  Greff.
\newblock Conditional object-centric learning from video.
\newblock In {\em International Conference on Learning Representations}, 2022.

\bibitem{Kosiorek2018SQAIR}
Adam Kosiorek, Hyunjik Kim, Yee~Whye Teh, and Ingmar Posner.
\newblock {Sequential Attend, Infer, Repeat: Generative Modelling of Moving
  Objects}.
\newblock In {\em Advances in Neural Information Processing Systems}, 2018.

\bibitem{lai2020mast}
Zihang Lai, Erika Lu, and Weidi Xie.
\newblock Mast: A memory-augmented self-supervised tracker.
\newblock In {\em Proceedings of the IEEE/CVF Conference on Computer Vision and
  Pattern Recognition}, pages 6479--6488, 2020.

\bibitem{liu2021retrieval}
Shangqing Liu, Yu Chen, Xiaofei Xie, Jing~Kai Siow, and Yang Liu.
\newblock Retrieval-augmented generation for code summarization via hybrid gnn.
\newblock In {\em International Conference on Learning Representations}, 2021.

\bibitem{liu2023causal}
Yuejiang Liu, Alexandre Alahi, Chris Russell, Max Horn, Dominik Zietlow,
  Bernhard Sch{\"o}lkopf, and Francesco Locatello.
\newblock Causal triplet: An open challenge for intervention-centric causal
  representation learning.
\newblock 2023.

\bibitem{locatello2020object}
Francesco Locatello, Dirk Weissenborn, Thomas Unterthiner, Aravindh Mahendran,
  Georg Heigold, Jakob Uszkoreit, Alexey Dosovitskiy, and Thomas Kipf.
\newblock Object-centric learning with slot attention.
\newblock {\em Advances in Neural Information Processing Systems},
  33:11525--11538, 2020.

\bibitem{lu2020video}
Xiankai Lu, Wenguan Wang, Martin Danelljan, Tianfei Zhou, Jianbing Shen, and
  Luc Van~Gool.
\newblock Video object segmentation with episodic graph memory networks.
\newblock In {\em Computer Vision--ECCV 2020: 16th European Conference,
  Glasgow, UK, August 23--28, 2020, Proceedings, Part III 16}, pages 661--679.
  Springer, 2020.

\bibitem{mambelli2022compositional}
Davide Mambelli, Frederik Tr{\"a}uble, Stefan Bauer, Bernhard Sch{\"o}lkopf,
  and Francesco Locatello.
\newblock Compositional multi-object reinforcement learning with linear
  relation networks.
\newblock {\em arXiv preprint arXiv:2201.13388}, 2022.

\bibitem{mansouri2022object}
Amin Mansouri, Jason Hartford, Kartik Ahuja, and Yoshua Bengio.
\newblock Object-centric causal representation learning.
\newblock In {\em NeurIPS 2022 Workshop on Symmetry and Geometry in Neural
  Representations}, 2022.

\bibitem{milan2016mot16}
Anton Milan, Laura Leal-Taix{\'e}, Ian Reid, Stefan Roth, and Konrad Schindler.
\newblock Mot16: A benchmark for multi-object tracking.
\newblock {\em arXiv preprint arXiv:1603.00831}, 2016.

\bibitem{niemeyer2021giraffe}
Michael Niemeyer and Andreas Geiger.
\newblock Giraffe: Representing scenes as compositional generative neural
  feature fields.
\newblock In {\em Proceedings of the IEEE/CVF Conference on Computer Vision and
  Pattern Recognition}, pages 11453--11464, 2021.

\bibitem{oh2019video}
Seoung~Wug Oh, Joon-Young Lee, Ning Xu, and Seon~Joo Kim.
\newblock Video object segmentation using space-time memory networks.
\newblock In {\em Proceedings of the IEEE/CVF International Conference on
  Computer Vision}, pages 9226--9235, 2019.

\bibitem{pei2019memory}
Wenjie Pei, Jiyuan Zhang, Xiangrong Wang, Lei Ke, Xiaoyong Shen, and Yu-Wing
  Tai.
\newblock Memory-attended recurrent network for video captioning.
\newblock In {\em Proceedings of the IEEE/CVF Conference on Computer Vision and
  Pattern Recognition}, pages 8347--8356, 2019.

\bibitem{pylyshyn1989role}
Zenon Pylyshyn.
\newblock The role of location indexes in spatial perception: A sketch of the
  finst spatial-index model.
\newblock {\em Cognition}, 32(1):65--97, 1989.

\bibitem{radford2019language}
Alec Radford, Jeffrey Wu, Rewon Child, David Luan, Dario Amodei, Ilya
  Sutskever, et~al.
\newblock Language models are unsupervised multitask learners.
\newblock {\em OpenAI blog}, 1(8):9, 2019.

\bibitem{rahaman2021dynamic}
Nasim Rahaman, Muhammad~Waleed Gondal, Shruti Joshi, Peter Gehler, Yoshua
  Bengio, Francesco Locatello, and Bernhard Sch{\"o}lkopf.
\newblock Dynamic inference with neural interpreters.
\newblock {\em Advances in Neural Information Processing Systems},
  34:10985--10998, 2021.

\bibitem{ristani2016performance}
Ergys Ristani, Francesco Solera, Roger Zou, Rita Cucchiara, and Carlo Tomasi.
\newblock Performance measures and a data set for multi-target, multi-camera
  tracking.
\newblock In {\em Computer Vision--ECCV 2016 Workshops: Amsterdam, The
  Netherlands, October 8-10 and 15-16, 2016, Proceedings, Part II}, pages
  17--35. Springer, 2016.

\bibitem{ruder2016overview}
Sebastian Ruder.
\newblock An overview of gradient descent optimization algorithms.
\newblock {\em arXiv preprint arXiv:1609.04747}, 2016.

\bibitem{seitzer2022bridging}
Maximilian Seitzer, Max Horn, Andrii Zadaianchuk, Dominik Zietlow, Tianjun
  Xiao, Carl-Johann Simon-Gabriel, Tong He, Zheng Zhang, Bernhard
  Sch{\"o}lkopf, Thomas Brox, et~al.
\newblock Bridging the gap to real-world object-centric learning.
\newblock {\em arXiv preprint arXiv:2209.14860}, 2022.

\bibitem{singh2022illiterate}
Gautam Singh, Fei Deng, and Sungjin Ahn.
\newblock Illiterate {DALL}-e learns to compose.
\newblock In {\em International Conference on Learning Representations}, 2022.

\bibitem{singh2022simple}
Gautam Singh, Yi-Fu Wu, and Sungjin Ahn.
\newblock Simple unsupervised object-centric learning for complex and
  naturalistic videos.
\newblock {\em arXiv preprint arXiv:2205.14065}, 2022.

\bibitem{stelzner2021decomposing}
Karl Stelzner, Kristian Kersting, and Adam~R Kosiorek.
\newblock Decomposing 3d scenes into objects via unsupervised volume
  segmentation.
\newblock {\em arXiv preprint arXiv:2104.01148}, 2021.

\bibitem{tangemann2021unsupervised}
Matthias Tangemann, Steffen Schneider, Julius Von~K{\"u}gelgen, Francesco
  Locatello, Peter Gehler, Thomas Brox, Matthias K{\"u}mmerer, Matthias Bethge,
  and Bernhard Sch{\"o}lkopf.
\newblock Unsupervised object learning via common fate.
\newblock {\em arXiv preprint arXiv:2110.06562}, 2021.

\bibitem{trauble2022discrete}
Frederik Tr{\"a}uble, Anirudh Goyal, Nasim Rahaman, Michael Mozer, Kenji
  Kawaguchi, Yoshua Bengio, and Bernhard Sch{\"o}lkopf.
\newblock Discrete key-value bottleneck.
\newblock {\em arXiv preprint arXiv:2207.11240}, 2022.

\bibitem{vaswani2017attention}
Ashish Vaswani, Noam Shazeer, Niki Parmar, Jakob Uszkoreit, Llion Jones,
  Aidan~N Gomez, {\L}ukasz Kaiser, and Illia Polosukhin.
\newblock Attention is all you need.
\newblock {\em Advances in neural information processing systems}, 30, 2017.

\bibitem{von2020towards}
Julius von K{\"u}gelgen, Ivan Ustyuzhaninov, Peter Gehler, Matthias Bethge, and
  Bernhard Sch{\"o}lkopf.
\newblock Towards causal generative scene models via competition of experts.
\newblock {\em arXiv preprint arXiv:2004.12906}, 2020.

\bibitem{wang2019learning}
Xiaolong Wang, Allan Jabri, and Alexei~A Efros.
\newblock Learning correspondence from the cycle-consistency of time.
\newblock In {\em Proceedings of the IEEE/CVF Conference on Computer Vision and
  Pattern Recognition}, pages 2566--2576, 2019.

\bibitem{wu2019long}
Chao-Yuan Wu, Christoph Feichtenhofer, Haoqi Fan, Kaiming He, Philipp
  Krahenbuhl, and Ross Girshick.
\newblock Long-term feature banks for detailed video understanding.
\newblock In {\em Proceedings of the IEEE/CVF Conference on Computer Vision and
  Pattern Recognition}, pages 284--293, 2019.

\bibitem{wu2022slotformer}
Ziyi Wu, Nikita Dvornik, Klaus Greff, Thomas Kipf, and Animesh Garg.
\newblock Slotformer: Unsupervised visual dynamics simulation with
  object-centric models.
\newblock {\em arXiv preprint arXiv:2210.05861}, 2022.

\bibitem{yang2018learning}
Tianyu Yang and Antoni~B Chan.
\newblock Learning dynamic memory networks for object tracking.
\newblock In {\em Proceedings of the European conference on computer vision
  (ECCV)}, pages 152--167, 2018.

\bibitem{yoon2023investigation}
Jaesik Yoon, Yi-Fu Wu, Heechul Bae, and Sungjin Ahn.
\newblock An investigation into pre-training object-centric representations for
  reinforcement learning.
\newblock {\em arXiv preprint arXiv:2302.04419}, 2023.

\bibitem{zeng2022motr}
Fangao Zeng, Bin Dong, Yuang Zhang, Tiancai Wang, Xiangyu Zhang, and Yichen
  Wei.
\newblock Motr: End-to-end multiple-object tracking with transformer.
\newblock In {\em Computer Vision--ECCV 2022: 17th European Conference, Tel
  Aviv, Israel, October 23--27, 2022, Proceedings, Part XXVII}, pages 659--675.
  Springer, 2022.

\bibitem{zhang2019robust}
Wenwei Zhang, Hui Zhou, Shuyang Sun, Zhe Wang, Jianping Shi, and Chen~Change
  Loy.
\newblock Robust multi-modality multi-object tracking.
\newblock In {\em Proceedings of the IEEE/CVF International Conference on
  Computer Vision}, pages 2365--2374, 2019.

\bibitem{zhao2021modelling}
Zixu Zhao, Yueming Jin, and Pheng-Ann Heng.
\newblock Modelling neighbor relation in joint space-time graph for video
  correspondence learning.
\newblock In {\em Proceedings of the IEEE/CVF International Conference on
  Computer Vision}, pages 9960--9969, 2021.

\bibitem{zhou2020tracking}
Xingyi Zhou, Vladlen Koltun, and Philipp Kr{\"a}henb{\"u}hl.
\newblock Tracking objects as points.
\newblock In {\em Computer Vision--ECCV 2020: 16th European Conference,
  Glasgow, UK, August 23--28, 2020, Proceedings, Part IV}, pages 474--490.
  Springer, 2020.

\bibitem{zou2023segment}
Xueyan Zou, Jianwei Yang, Hao Zhang, Feng Li, Linjie Li, Jianfeng Gao, and
  Yong~Jae Lee.
\newblock Segment everything everywhere all at once.
\newblock {\em arXiv preprint arXiv:2304.06718}, 2023.

\end{thebibliography}
}

\clearpage
\appendix
\renewcommand{\thesection}{\Alph{section}.\arabic{section}}
\setcounter{section}{0}

\begin{appendices}
\section{EM-inspired Loss Formulation}
In this section, we illustrate the formulation of our EM-inspired loss with more details. First, let's start from the formulation of the Expectation-Maximization (EM) algorithm.
Given data $X$, parameters $\theta$ and a set of unobserved latent data $Z$, we intend to estimate the following data likelihood:
\begin{align}
    p(X|\theta) =\int p(X|Z, \theta) p(Z|\theta) dZ =\mathbb{E}_{Z\sim p(\cdot|\theta)}[p(X|Z, \theta)].
\end{align}
However, $p(Z|\theta)$ is usually intractable in practice, therefore the EM algorithm proposes to instead optimize $\theta$ with the following two-step iterative process:

\begin{enumerate}
    \item Expectation step: Given $\theta^{(t)}$ as the estimation value at iteration step $t$, we compute the log-likelihood $Q(\theta|\theta^{(t)})=\mathbb{E}_{Z\sim p(\cdot|\theta^{(t)}, X)}[\log p(X, Z|\theta)]$.
    \item Maximization step: We obtain the next value by $\theta^{(t+1)}=\arg\min_\theta Q(\theta|\theta^{(t)})$.
\end{enumerate}
Suppose $Z$ is discrete, the E-step can be re-written as:
\begin{align}
    Q(\theta|\theta^{(t)}) &= \mathbb{E}_{Z\sim p(\cdot|\theta^{(t)}, X)}[\log p(X, Z|\theta)]\\
    &=\sum_Z p(Z|\theta^{(t)}, X) \log p(X, Z|\theta)
\end{align}
In our work, we see $\mathcal{I}$ as the unobserved data $Z$, and therefore $\mathcal{I}_{ij}$ as a sample from the distribution $p(Z|\theta^{(t)}, X)$. 
Next, we formulate $\log p(X, Z|\theta)$ with three likelihood functions by design. Let $\text{Dec}(\cdot)$ be the mask decoder shared by slots and memory, the memory representation $\mathcal{M}$ and slot representation $\mathcal{S}$ be two $\theta$-parameterized functions of $\mathcal{I}$ (i.e. $Z$) and $X$.

\begin{enumerate}
    \item $p_1(X, Z|\theta) = \text{Dec}(\mathcal{M})^{\text{Dec}(\mathcal{S})}$ 
    \item $p_2(X, Z|\theta)=\mathcal{N}(\text{Dec}(\mathcal{M});\text{Dec}(\mathcal{S}), I)$
    \item $p_3(X, Z|\theta)=\mathcal{N}(\mathcal{M};\mathcal{S}, I)$
\end{enumerate}

\noindent where $I$ is the identity matrix. Next, let

\begin{equation}
    p(X, Z|\theta) \propto p_1(X, Z|\theta)^{\lambda_1} p_2(X, Z|\theta)^{2\lambda_2} p_3(X, Z|\theta)^{2\lambda_3}
\end{equation}

\noindent and we will have:
\begin{align}
    \log p(X, Z|\theta) = & \lambda_1 \text{Dec}(\mathcal{S}) \log \text{Dec}(\mathcal{M}) \\
    & - \lambda_2 ||\text{Dec}(\mathcal{S}) - \text{Dec}(\mathcal{M})||^2 \\
    & - \lambda_3 ||\mathcal{S} - \mathcal{M}||^2 + C,
\end{align}

\noindent with $C$ being negligible constant.

By taking negative and expanding $\mathcal{S}$ to $\mathcal{S}^i$ and $\mathcal{M}$ to $\mathcal{M}^j$, we have the equivalence between $\log p(X, Z|\theta)$ and Equation (5), i.e. $\mathcal{L}_{assign}$, in the main text. As a result, Equation (7) is equivalent to $Q(\theta|\theta^{(t)})$.
To simplify the computation of $\arg\min_\theta Q(\theta|\theta^{(t)})$ for the M-step, we use stochastic gradient descent to approximate $\theta^{(t+1)} = \theta^{(t)} - \alpha \nabla Q(\theta|\theta^{(t)})$.
\section{Hyper-parameter Selection}
Table~\ref{tab:param} summarizes the hyper-parameter selection for OC-MOT on CATER and FISHBOWL datasets. We ablate the selection of those hyper-parameters and show the best choice in bold. Specifically, we analyzed the effect of memory length and MHA block number in the main text.
\begin{table}[!ht]
	\centering
\resizebox{0.46\textwidth}{!}{
\begin{tabular}{ccc}
\toprule[1.5pt]
Method     & CATER       & FISHBOWL  \\\hline
 $\tau_{out}$ &  \textbf{5}, 7, 9 & 5  \\
$\tau_{iou}$ &  \textbf{0.9}, 0.8, 0.7 & 0.9  \\
$\lambda_1, \lambda_2, \lambda_3$  &$1, 0.1, 0$   & $1, 0, 1$  \\  
Memory Length $T_{max}$   & \textbf{6},10,20,32   & 6  \\ 
Slot Number $N$ &  11 & 24  \\
Buffer Number $M$ &  12, \textbf{15}, 20 & 40  \\
 MHA Block Number & 1, \textbf{2}   &  2 \\
\bottomrule[1.5pt]
\end{tabular}
}
\vspace{+0.05in}
\caption{\textbf{Hyper-parameter selection for OC-MOT}. The best selections are marked in bold. }
	\label{tab:param}
\end{table} 

\section{OC Grouping with Partial Labels}
\begin{figure}[t]
	\centering
	\includegraphics[width=0.48\textwidth]{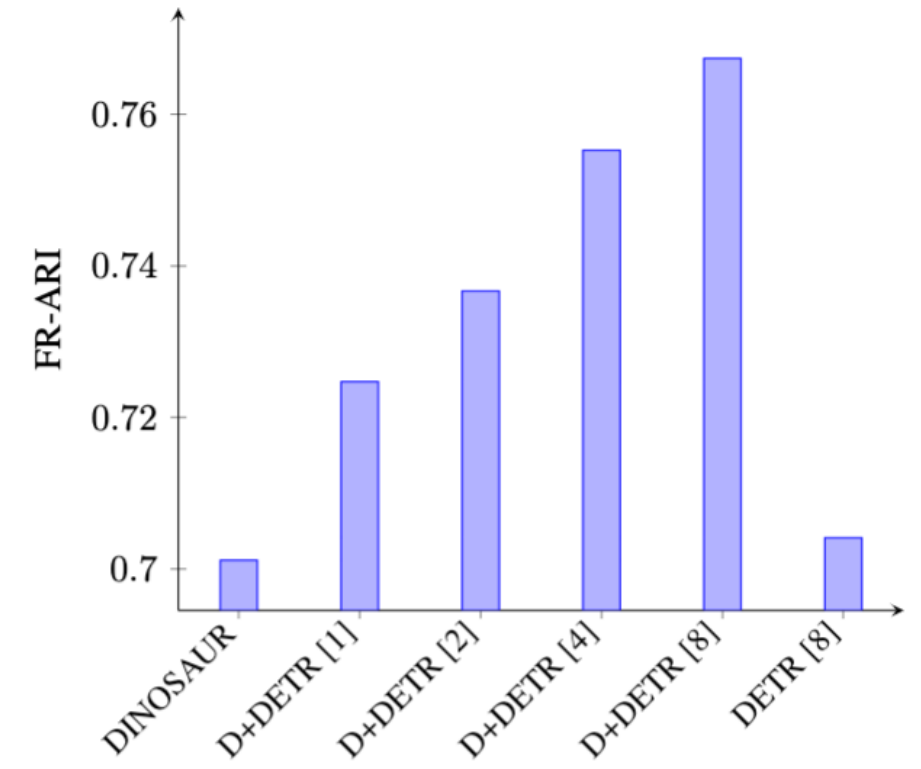}
\caption{\textbf{Effect of partial labeling on OC grouping on FISHBOWL.} Comparison of FG-ARI for training fully unsupervised (DINOSAUR), partially supervised (D+DETR [\#annotated frames]) and fully supervised on 8 frames per video (DETR [8]).}
\label{fig:ablation}
\end{figure}

To understand how labels and unsupervised training objectives can be combined in a synergistic fashion, we performed a series of ablations where we decrease the number of annotated frames gradually.  Generally, it is to be expected, that the performance continues to increase as more labels are added, yet this would be linked to higher labeling effort/cost in real-world applications. Thus, in a practical setting, it is desired to find the point of diminishing returns, where the rate of performance increase declines as more data is added. As these experiments are independent from the memory module, they were performed by training a DINOSAUR model on a per-frame basis on the FISHBOWL dataset.
In order to reduce variance in this comparison, all models are fine-tuned based on the same checkpoint of a vanilla trained DINOSAUR model~\cite{seitzer2022bridging} and use the same subset of frames. 
As shown in Figure~\ref{fig:ablation}, both the purely unsupervised model (DINOSAUR) as well as the fully supervised version of the model on a subset of 8 frames (DETR [8]) show comparable results.  Nevertheless, by adding some annotated frames the performance increases significantly as ambiguity about part-whole relationships is resolved by the labels. The relative performance increase of doubling the number of annotated labels is on average $1.92$\%. Further, we see indication of diminishing returns when increasing the number of frames from 4 to 8 as the performance only improves by $1.6$\% in this case. Overall, this shows that adding a few annotated examples to the data significantly improves the performance of the unsupervised DINOSAUR model and greatly outperforms a model that was exclusively trained on the annotated data.

\begin{table}[!ht]
	\centering
\resizebox{0.45\textwidth}{!}{
\begin{tabular}{lccc}
\toprule[1.5pt]
Method   & IDF1 $\uparrow$   & MOTA $\uparrow$       & IDS $\downarrow$  \\\hline
SAVi (RGB Recon.)  & 46.9\% & 32.3\%   & 12504 \\  
SAVi (Optical Flow) & 53.2\% & 34.9\% &  15394  \\  
\textbf{OC-MOT} &  \textbf{77.9\%} & \textbf{70.3\%}  & \textbf{5898}  \\
\bottomrule[1.5pt]
\end{tabular}
}
\caption{\textbf{Comparison with video object-centric models on FISHBOWL}. The SAVi models are trained with RGB reconstruction and optical flow reconstruction, respectively.}
	\label{tab:results}
\end{table}

\section{SAVi Baseline Analysis}
We reported the MOT results of SAVi on FISHBOWL dataset in the main paper. To handle the grouping problems in complicated scenes, such as over-segmentation (part-whole issue),  we trained the SAVi baseline with supervised DETR-style loss (with 6.25\% detection labels) on top of the self-supervised RGB reconstruction loss. 
 Another option to run the SAVi reconstruction loss is to reconstruct the optical flow and this setting is reported to work better on complicated scenes. As shown in Table~\ref{tab:results}, SAVi with optical flow achieves slightly better IDF1 and MOTA, and even more ID-Switches. The performance still has a large gap with our proposed OC-MOT. In the deep dive analysis, we noticed SAVi with optical flow reconstruction improves on the boundary accuracy for the segmentation masks, but shows similar issues on ID Switch and over-segmentation.

\section{Extension to Real-world Videos.}

\begin{figure}[t]
	\centering
	\includegraphics[width=0.48\textwidth]{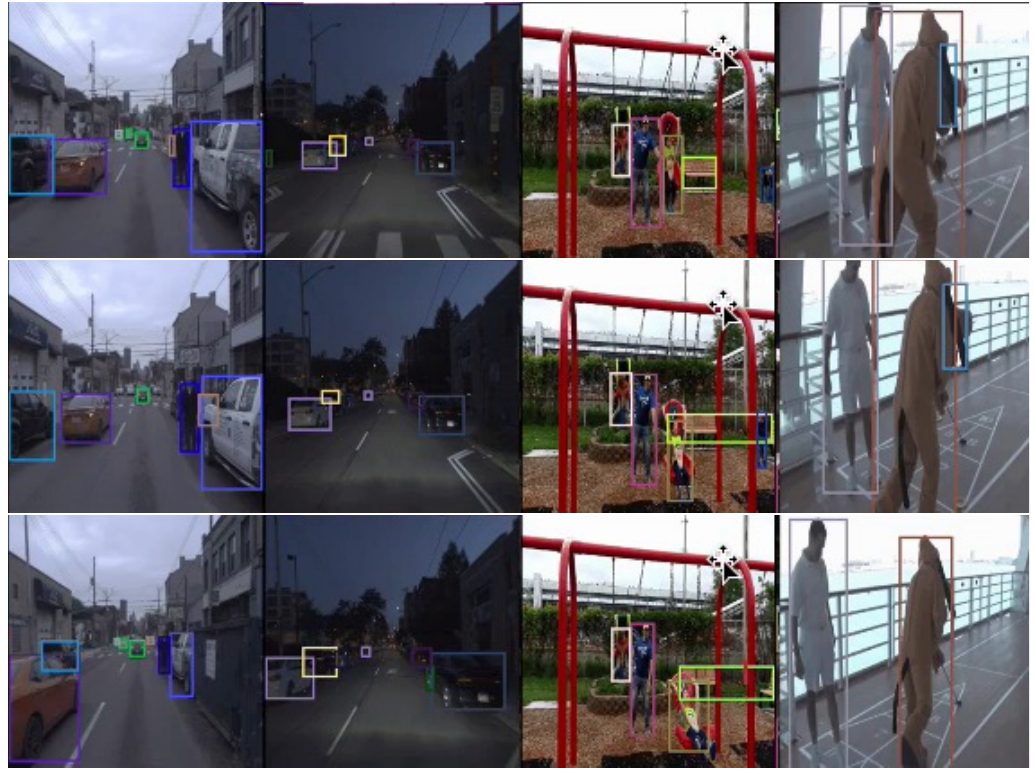}
\caption{\textbf{MOT results of OC-MOT on TAO dataset}. Only the tracklets of persons and cars are visualized.}
\label{fig:reald}
\end{figure}

In the main paper, we do not report metrics on standard benchmarks such as MOT17 that  heavily reflects the detection performance rather than object association performance, as majority of the bounding boxes can be correctly linked with spatial locations. Therefore it's not a good fit for evaluating our method because current un/weakly-supervised OC models are less capable of producing comparable results with SOTA supervised object detectors. We discussed this as limitation in Section 4.1 on KITTI, which is generalizable to MOT17. Instead, \textbf{one highlight of the work is our novel framework to learn object association in a self-supervised manner}. This is \textit{agnostic to the detection module}. We believe this contribution is quite interesting and novel in MOT community. To further apply OC-MOT to real-world videos, we replaced the object-centric model with SEEM~\cite{zou2023segment} that can accurately segment objects in real worlds. We trained OC-MOT with self-supervised loss on TAO (track any object) dataset and observed quite good tracking performance. In Figure~\ref{fig:reald} , we visualize the tracklets of persons and cars. Overall, OC-MOT performs strong object association without ID labels as long as the detection model can provide good object representations that contain enough information for tracking such as appearances and locations.

\end{appendices}

\end{document}